\documentclass[letterpaper, 10 pt, conference]{ieeeconf}  

\IEEEoverridecommandlockouts                              

\usepackage[T1]{fontenc}

\usepackage{amsmath} 
\usepackage{amssymb}  
\usepackage{graphicx}
\usepackage{tikz}
\usepackage{booktabs}
\usepackage{multirow}
\usepackage{multicol}
\usepackage{placeins}
\usepackage{hyperref}
\newcommand{\Norm}[1]{\left\lVert#1\right\rVert}
\newcommand{\loss}{\mathcal{L}}

\title{\LARGE \bf
Revisiting the Adversarial Robustness-Accuracy Tradeoff in \\Robot Learning
}

\author{Mathias Lechner$^{1}$, Alexander Amini$^{1}$, Daniela Rus$^{1}$, Thomas A. Henzinger$^{2}$ 
\thanks{$^{1}$Massachusetts Institute of Technology, Computer Science and Artificial Intelligence Laboratory, Cambridge, MA, 02139 USA
         {\tt\small \{mlechner, amini, rus\}@mit.edu)}}%
\thanks{$^{2}$Institute of Science and Technology Austria (IST Austria)
       {\tt\small tah@ist.ac.at)}}
}

\begin{document}

\maketitle
\thispagestyle{empty}
\pagestyle{empty}

\begin{abstract}
Adversarial training (i.e., training on adversarially perturbed input data) is a well-studied method for making neural networks robust to potential adversarial attacks during inference.
However, the improved robustness does not come for free but rather is accompanied by a decrease in overall model accuracy and performance.
Recent work has shown that, in practical robot learning applications, the effects of adversarial training do not pose a fair trade-off but inflict a net loss when measured in holistic robot performance.
This work revisits the robustness-accuracy trade-off in robot learning by systematically analyzing if recent advances in robust training methods and theory in conjunction with adversarial robot learning, are capable of making adversarial training suitable for real-world robot applications.
We evaluate three different robot learning tasks ranging from autonomous driving in a high-fidelity environment amenable to sim-to-real deployment to mobile robot navigation and gesture recognition.
Our results demonstrate that, while these techniques make incremental improvements on the trade-off on a relative scale, the negative impact on the nominal accuracy caused by adversarial training still outweighs the improved robustness by an order of magnitude.
We conclude that although progress is happening, further advances in robust learning methods are necessary before they can benefit robot learning tasks in practice.
\end{abstract}

\section{Introduction}
Adversarial attacks are well-studied vulnerabilities of deep neural networks \cite{szegedy2013intriguing,goodfellow2014explaining}. These norm-bounded input perturbations make the network change its decision compared to the unaltered input and can have catastrophic impact in practical robotics applications. Critically, the adversarially altered inputs are barely distinguishable from the original input by humans. 
Most realistic-sized computer vision networks can be fooled by perturbations that change each pixel by a maximum of 4\% (i.e., a $l_\infty$-norm less or equal to 8) while being barely noticeable by humans.

Adversarial robustness is an important consideration in the development of robotic applications, as it ensures that the robot's behavior remains consistent and predictable in the presence of perturbations or attacks. In the real world, robots must be able to operate in a variety of environments and under a wide range of conditions, some of which may be outside of their training data or beyond their control. Adversarial robustness allows robots to continue functioning effectively even when faced with such challenges, improving their reliability and safety in real-world applications. Additionally, as robots become more integrated into society and are given greater autonomy, it becomes increasingly important to ensure that they are not susceptible to manipulation or exploitation by malicious actors. Adversarial robustness helps to protect against such threats and ensure that robots can be trusted to behave in a predictable and responsible manner.

\begin{figure}
    \centering
    \includegraphics[width=\linewidth]{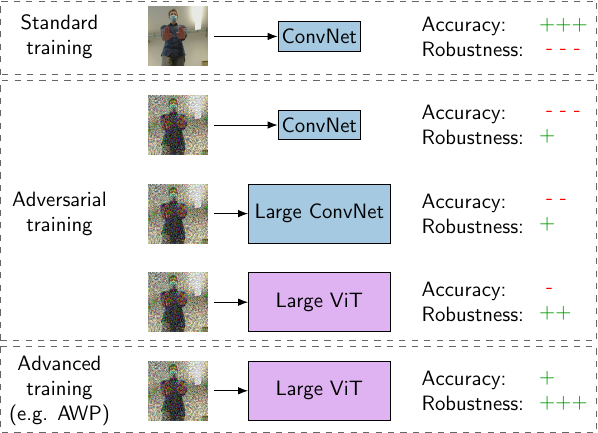}
    \caption{High-level summary of our results. Adversarial training improves robustness at the cost of significantly reduced accuracy. We show that methods to counteract this decrease in accuracy are most effective when multiple approaches are combined, i.e., an overparametrized network, a vision transformer neural architecture, and advanced adversarial training procedures.}
    \label{fig:teaser}
\end{figure}

Robust learning aims to tackle the problem by training networks that are immune to adversarial or other types of attacks \cite{wald1945statistical,huber1964robust,atkeson1997robot,xu2009robustness,madry2017towards,song2018improving,biggio2012poisoning,konstantinov2019robust}. One of the most dominant approaches for training robust models is adversarial training which adds adversarial perturbations to the training data online during and throughout the learning procedure \cite{wald1945statistical,madry2017towards}.
Adversarial training methods improve the test-time robustness on adversarial examples at the critical cost of lower nominal accuracy \cite{raghunathan2019adversarial,zhang2018efficient,tsipras2018robustness}. 
For instance, the advanced adversarial training algorithm of \cite{zhang2019theoretically}, which won the NeurIPS 2018 Adversarial Vision Challenge, yielded a robust network with an accuracy of 89\% on the CIFAR-10 dataset. In contrast, standard training algorithms can easily produce non-robust networks with an accuracy above 96\% on this dataset \cite{zagoruyko2016wide}. 
This dilemma of choosing between an accurate but vulnerable and a robust but less accurate model is known as the robustness-accuracy trade-off \cite{tsipras2018robustness,zhang2019theoretically,raghunathan2019adversarial}.

Recent work \cite{lechner2021adversarial} has investigated this trade-off specifically in the context of robot learning applications where both accuracy and robustness are critical as the system is ultimately deployed into physical, safety-critical environments. The authors observed that this trade-off is not fair trade but poses a net loss when evaluating the robots' overall performance and concluded that adversarial training is not ready for robot learning.
However, recent work has shown that multiple factors (e.g., model size, choice of the activation function, adversarial training procedure) contribute to the reduction in accuracy of robot learning methods \cite{rebuffi2021data,bubeck2021law,zhou2022understanding,singla2021low,pang2021bag,wu2020adversarial}.
In particular, these works underline that larger models are necessary for robustly fitting the training data \cite{bubeck2021universal,bubeck2021law}. Moreover, they emphasize that a more careful selection of the neural network architecture and hyperparameters is needed when replacing standard training with adversarial training methods \cite{pang2021bag,zhou2022understanding}. However, there remains a critically important and open question on if these recent advances are sufficient to quell the costs of adversarial training for robotics.  

In this work, we assess whether the conclusion of \cite{lechner2021adversarial} that adversarial training is not ready for robot learning remains true or is challenged by these recent advances in the field.
In particular, we evaluate if overparametrized models \cite{bubeck2021law}, vision transformers \cite{zhou2022understanding,paul2022vision}, smooth curvature activation functions \cite{singla2021low}, more careful hyperparameter selection \cite{pang2021bag}, and advanced adversarial training methods \cite{wu2020adversarial} can provide acceptable accuracy and robustness on three robot learning and autonomous driving tasks.

Our results show that, although the techniques listed above pose a significant improvement in the robustness-accuracy gap, the negative impact on the nominal accuracy from adversarial training still outweighs the benefits of the induced robustness. 
Specifically, while the methods from the literature make single digits improvements on the robustness-accuracy Pareto front, i.e., improving both accuracy and robustness, the negative side-effects of adversarial training methods still outweigh these advances by an order of magnitude.
Nonetheless, we observed the trend that combining multiple individually introduced robustness enhancement methods provided the most promising future path toward closing the robustness-accuracy gap, e.g., as outlined in Figure \ref{fig:teaser}. 

We summarize our contributions as:
\begin{itemize}
	\item We evaluate five advancements in robust learning methods in three different real-world robotic applications 
(456 models tested in total) for their suitability in closing the robustness-accuracy tradeoff gap in robot learning tasks in practice.
	\item We provide strong empirical evidence that, while robustness can be improved by the methods from literature, the negative effect on the nominal accuracy of adversarial training still outweighs the improvements of these methods by an order of magnitude.
    \item Our results show that adversarial training is most effective when multiple individual robust learning approaches are combined. This suggests that the most promising path to closing the robustness-accuracy gap entirely in the future is the integration of multiple independent approaches for enhancing robustness.
\end{itemize}

The remainder of this paper is structured as follows. In section \ref{sec:adv}, we recapitulate robustness of neural networks, adversarial training, and the robustness-accuracy trade-off. In section \ref{sec:methods}, we describe related work on improving the robustness of neural networks and avoiding the reduced clean accuracy of adversarial training. Finally, in section \ref{sec:experiments}, we experimentally evaluate these improvements on three robot learning tasks.

\section{Background and Related Work}\label{sec:adv}
A neural network is a function $f_\theta: \mathcal{X} \rightarrow \mathcal{Y}$ parameterized by $\theta$.
In supervised learning, the training objective is to fit the function to a given dataset in the form of $\{(x_1,y_1),\dots (x_n,y_n)\}$ assumed to be i.i.d. sampled from a probability distribution over $\mathcal{X} \times \mathcal{Y}$. 
This fitting process is done via empirical risk minimization (ERM) that minimizes
\begin{equation}\label{eq:erm}
\frac{1}{n} \sum_{i=1}^{n} \loss(f_\theta(x_i),y_i)
\end{equation} 
via stochastic gradient descent. The differentiable loss function $\loss: \mathcal{Y}\times \mathcal{Y} \rightarrow \mathbb{R}$  characterizes how well the network's prediction $f_\theta(x_i)$ matches the ground truth label $y_i$.

An adversarial attack is a sample $(x_i,y_i)$ from the data distribution and a corresponding attack vector $\mu$ with $\Norm{\mu} \leq \varepsilon$ such that $f(x_i) \neq f(x_i+\mu)$ with $\varepsilon$ being a threshold. 
For image data, $L_\infty$ thresholds $\delta\leq 8$ are usually not recognizable or appear as noise for human observers.
It has been shown that most neural networks, irrespective of network types, input domains, or learning setting, are susceptible to adversarial attacks \cite{goodfellow2014explaining,brown2017adversarial,uesato2018adversarial,athalye2018obfuscated,schonherr2018adversarial,giacobbe2020many,bai2021transformers}.

Typical norms used in adversarial attacks are the $\ell_1$, $\ell_2$, and the $\ell_\infty$ norm. In this work, we focus on the $\ell_\infty$ norm.
A network is robust on a given sample if no such attack $\mu$ exists within a threshold $\varepsilon$. 
The robust accuracy is the standard metric for measuring the robustness of a network aggregated over an entire dataset $\{(x_1,y_1),\dots (x_n,y_n)\}$ by counting the ratio of correctly classified samples that are also robust. 

In practice, deciding whether a network is robust for a sample is an NP-hard problem \cite{katz2017reluplex,salzer2021reachability,henzinger2021scalable} and, therefore, cannot be computed for typically sized networks in a reasonable time.
Instead, the robustness of networks is often studied with respect to empirical gradient and black-box-based attack methods. 
The fast gradient sign method (FGSM) \cite{goodfellow2014explaining} computes an attack by
\begin{equation}
	\mu = \varepsilon\ \text{sign}\Big(\frac{\partial\loss(f_\theta(x_i),y_i)}{\partial x_i}\Big).
\end{equation}
Despite its simplicity, adversarial training often uses the FGSM method due to its speed.
The iterative fast gradient sign method (I-FGSM) \cite{kurakin2016adversarial} is a more sophisticated generalization of the FSGM. It computes an attack iteratively in $k$ steps starting from $\mu_0 = \mathbf{0}$ and updating it by
\begin{equation}
\mu_i = \frac{\varepsilon}{k}\ \text{sign}\Big(\frac{\partial\loss(f_\theta(x_i+\mu_{i-1}),y_i)}{\partial x_i}\Big).
\end{equation}

DeepFool \cite{moosavi2016deepfool}, the C\&W method \cite{carlini2017towards}, and projected gradient descent \cite{madry2017towards} are other common iterative attack methods that are used for evaluating robustness but are too computationally expensive to incorporate in adversarial training. DeepFool \cite{moosavi2016deepfool} linearizes the network in each iteration of updating $\mu_i$. Projected gradient descent \cite{madry2017towards} applies unconstrained gradient descent but divides each $\mu_i$ by its norm and multiplies the results with $\varepsilon$ to project it back into the given threshold.
The C\&W method \cite{carlini2017towards} avoids such projection by parametrizing the attack vector $\mu$ by another variable and a transformation that already normalizes the attack to stay within a given threshold.
It has been experimentally shown that any network of non-trivial size is, at least in parts, vulnerable to such attacks \cite{madry2017towards}. 

Robust learning methods aim to train networks that are robust \cite{wald1945statistical,huber1964robust,atkeson1997robot,xu2009robustness,madry2017towards,song2018improving,biggio2012poisoning}.
One of the most common robust learning methods is adversarial training which changes the standard ERM objective to the min-max objective
\begin{equation}\label{eq:at}
\frac{1}{n} \sum_{i=1}^{n} \max_{\mu: \Norm{\mu}\leq \varepsilon} \loss(f_\theta(x_i+\mu),y_i),
\end{equation}
where $\varepsilon>0$ is some attack budget controlling how much each input can be perturbed. Due to the computation overhead by this training objective, fast attack-generating methods are typically used for computing the $\max$ in Equation \ref{eq:at}, e.g., the FGSM or I-FGSM.

Alternative approaches to adversarial training make minor modifications to the objective term in Equation \ref{eq:at}. For instance, the TRADES algorithm \cite{zhang2019theoretically} replaces the label $y_i$ in Equation \ref{eq:at} with the network's prediction of the original input, i.e., $f_\theta(x_i)$, and optimizes a joint objective of the standard ERM term and the robustness term.  
The approach of \cite{shafahi2019forfree} removes the overhead imposed by the maximization step in Equation \ref{eq:at} by pre-computing $\mu$ in the previous gradient descent step. Although such pre-computed $\mu$ can become inaccurate, i.e., stale, \cite{shafahi2019forfree} showed that it improves robustness in practice. 
Adversarial weight perturbation (AWP) \cite{wu2020adversarial} improves the generalization of adversarially trained networks by injecting adversarial noise into the weights of the network and smoothing the loss surface. Data augmentation applied to adversarial training has also been shown to positively affect the robustness, and the generalization of neural networks \cite{rebuffi2021data}. The work of \cite{lamb2022interpolated} has shown that the negative impact of adversarial training on the clean accuracy of a network can be further reduced by combining it with advanced data augmentation techniques such as MixUp \cite{zhang2018mixup}.

The major limitation of adversarial training methods is that they negatively affect the network's standard accuracy (or other performance metrics). 
For example, medium-sized networks achieve an accuracy of 96\% on the CIFAR-10 dataset when trained with standard ERM \cite{zagoruyko2016wide}. However, in \cite{zhang2019theoretically} the best-performing network trained with the TRADES algorithm could only achieve a standard accuracy of 89\% on this dataset.
This phenomenon of an antagonistic relation between accuracy and robustness was first studied in \cite{tsipras2018robustness} and is known as the accuracy-robustness trade-off. 
The trade-off was studied in the context of robot learning in \cite{lechner2021adversarial} by investigating whether the gained robustness is worth the reduction in nominal accuracy in real-world robotic tasks. The authors observed that the adversarially trained networks resulted in a worse robot performance than by using a network trained in the standard way. 

The concept of adversarial training and the min-max objective of robust learning has been adopted for other task-specific types of specifications, such as safety. For example, \cite{lechner2021adversarial} has introduced safety-domain training by replacing the norm-bounded neighborhoods of labeled samples with arbitrary sets and corresponding labels, i.e., a min-max training objective over labeled sets.
Some modifications of the min-max objective have been studied in feedback systems with closed-loop safety and stability specifications. For instance, \cite{ChangRG19, lechner2021infinite,lechner2021stability} propose to learn a safety certificate via a learner-verifier framework where the maximization step is replaced by a verification module that provides formal guarantees on the certificate.

Adversarial training has also been studied as a regularizer for improving the generalization of neural networks.
In particular, \cite{herrmann2022pyramid} used mild adversarial attacks based on a hierarchical structure to improve the clean accuracy of vision transformer models \cite{dosovitskiy2020image}.
The work of \cite{duan2019robot} studied human adversaries to improve the performance in robotic object manipulation tasks.

\section{Methods}\label{sec:methods}
In this section we describe three directions from the literature that point to paths of how to improve robustness without sacrificing standard accuracy.

\subsection{Smooth activations and bag of tricks}

Recent work suggests that the common ReLU activation function, i.e., $\max\{0,x\}$, is not well suited for adversarial training methods \cite{singla2021low}.
Instead, the authors observed that activation functions with smooth curvatures provide better robustness at roughly the same standard accuracy. 
Specifically, the sigmoid-weighted linear unit (SiLU) activation function \cite{elfwing2018sigmoid}, i.e., $x\cdot \frac{1}{1+\exp(-x)}$, was highlighted as having a smooth second derivative and observed to improve robustness compared to alternative activations. We note that the SiLU activation was concurrently proposed as swish activation function in \cite{ramachandran2017searching}. 

The work of \cite{pang2021bag} investigated how hyperparameters of the learning process affect adversarial training compared to standard ERM. For example, the authors experiment with learning rate schedules, early stopping, and batch size, among other settings.
The authors observed that adversarial training benefits from a higher weight decay factor than standard training. 
Moreover, the authors confirmed that a smooth activation function improves robustness over the ReLU activation. 

\begin{figure}
	\centering
        \vspace{0.15cm}
	\begin{tikzpicture}
	\node at (0,1) [align=center] {Summer\\(in-distribution)};
	\node at (4,1)  [align=center]{Winter\\(in-distribution)};
	\node at (0,-1.2) [align=center] {Autumn\\(out-of-distribution)};
	\node at (4,-1.2)  [align=center]{Night\\(out-of-distribution)};
	\node at (0,0) {\includegraphics[width=3.4cm]{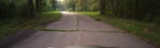}};
	\node at (4,0) {\includegraphics[width=3.4cm]{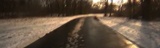}};
	\node at (0,-2.2) {\includegraphics[width=3.4cm]{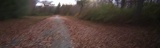}};
	\node at (4,-2.2) {\includegraphics[width=3.4cm]{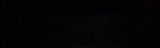}};
	\end{tikzpicture}
	\caption{Test conditions of our closed-loop driving experiment using a data-driven simulation environment \cite{amini2021vista}. The training data are collected in summer and winter conditions (separated from the testing data).}
	\label{fig:driving}
\end{figure}
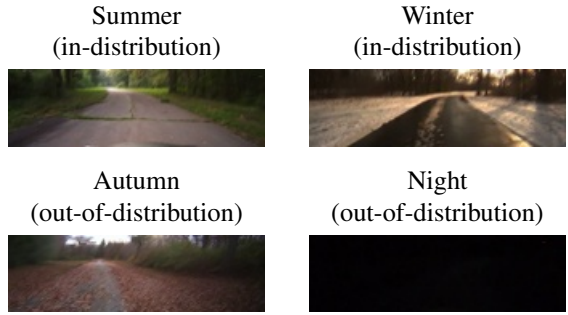

\subsection{Robustness requires overparametrization}
Theoretical contributions to the robustness-accuracy tradeoff recently discovered that overparametrization is necessary for smoothly fitting the training data \cite{bubeck2021universal}. While empirical results already suggested that the accuracy of larger models suffers less from adversarial training than for small models, the critical insight is that such large models are necessary. In particular, the authors proved that for a dataset of $n$ samples with $d$-dimensional features, a model with $n$ parameters can fit the training samples but cannot smoothly interpolate between them. Moreover, the authors show that a model needs at least $nd$ parameters to fit the training data and interpolate them smoothly.
The authors also demonstrated that contemporary models for standard datasets do not contain enough parameters with respect to their proven results.

\subsection{Vision transformers are more robust than CNNs}
The vision transformer (ViT) \cite{dosovitskiy2020image} is a powerful machine learning architecture that represents an image as a sequence of patches and processes this sequence using a self-attention mechanism \cite{vaswani2017attention}. 
Detailed experimental comparisons between vision transformer and convolutional neural networks suggest that ViTs are naturally more robust with respect to object occlusions and distributions shifts \cite{naseer2021intriguing}.
Concurrent work on comparing ViTs to CNNs with respect to adversarial attacks has found that vision transformers seem to be naturally more robust to adversarial attacks as well.

All advances on the robustness-accuracy tradeoff discussed above are either theoretical or were evaluated on static image classification tasks. 
Moreover, the methods are typically evaluated on research datasets such as CIFAR and ImageNet. While these datasets allow studying machine learning models' general performance, they significantly differ from real-world robot learning tasks. For example, the CIFAR datasets consist of very low-resolution images, i.e., 32-by-32 pixel, whereas robotic vision processing systems handle images with much higher resolution, e.g., 256-by-256 pixels in \cite{lechner2021adversarial}. Although the samples of the ImageNet dataset have a realistic image resolution, typical robot learning datasets consist of multiple orders of magnitude fewer samples than the ImageNet dataset.
Moreover, experiments on research datasets often report static test metrics, whereas learned robotic controllers are deployed in a closed-loop on a robot. 

The next section evaluates the methods described above on multiple real-world robot learning tasks, including open-loop training and closed-loop evaluation on an autonomous driving task. 

\section{Experiments}\label{sec:experiments}
In this section we study the advances in adversarial training methods on three robot learning tasks.

\subsection{End-to-end driving}
Our first experiment considers an autonomous driving task. In particular, a network is trained to predict the curvature of the road ahead of a car from images received at a camera that is mounted on top of the vehicle.
The training data is collected by a human driver who maneuvers the car around a test track. The networks are then trained on collected data using supervised learning. Finally, we deploy the networks in a closed-loop autonomous driving simulator. We use the VISTA simulation environment \cite{amini2021vista} for this purpose.

\begin{figure}
    \centering
    \includegraphics[width=\linewidth]{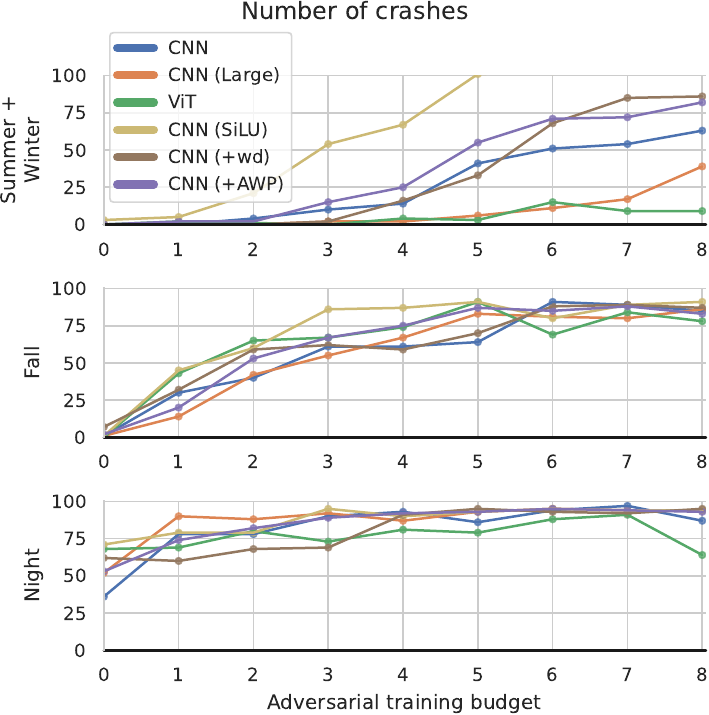}
    \caption{Number of crashes out of 100 simulation runs in each data setting (summer, winter, fall, night) with respect to varying the adversarial training budget. All models were trained in summer and winter conditions (on a different data split than the evaluations). The large CNN and the ViT model perform best under heavy adversarial training, but no adversarially trained model could handle distribution shifts, i.e., fall and night conditions.}
    \label{fig:vistaresults}
\end{figure}

We compare the performance of a baseline CNN with four variations. First, we compare with an enlarged variant of the baseline CNN to validate the necessity of overparametrization for robustness empirically. Next, we equip the baseline with the smoother SiLU activation and increase the weight decay (wd+). We also test the CNN trained with adversarial weight perturbation (AWP) \cite{wu2020adversarial} instead of training via the objective in Eq.~(\ref{eq:at}). Finally, we test a vision transformer model. 
The baseline model (CNN) consists of 440k, the enlarged model (CNN-large) of 7.7M, and the tested vision transformer (ViT) of 2.0M trainable parameters. The inputs of all architectures are 160-by-48 RGB images that are normalized per-image to have zero mean and unit standard deviation. 
The architecture details of the two convolutional networks are listed in the Appendix. Our vision transforms splits the input image into non-overlapping patches of 16-by-12 pixels, uses a latent dimension of 256, with 4 attention heads, 384 feed-forward dimensions, and 4 layers in total.
For the training, we use the Adam optimizer \cite{kingma2014adam} with a learning rate of 0.0003 and a batch size of 64. The weight decay is set to $10^{-5}$, except for the wd+ variant, which is trained with a decay factor of $5\cdot 10^{-5}$. We train all networks for a total of 900,000 steps.
We train all models with standard and adversarial training with increasing attack budget ($\varepsilon=0,1,\dots 8$) and I-FGSM as attack methods.

For each model and attack budget pair, we run a total of 400 simulations, split into 200 in-training distribution, and 200 out-of-training distribution condition runs. The in-training data were collected in summer and winter and were separated from the training data, i.e., there is no overlap between the training data and the evaluation data. The out-of-training data were collected in autumn and during the night, with no such condition present in the training data. The four conditions are visualized in Figure \ref{fig:driving}.
As an evaluation metric, we report the number of crashes during the simulation, i.e., when the vehicle leaves the road.

The top row in Figure \ref{fig:vistaresults} shows the crashes during the summer and winter simulations. The results show that the overparametrized model and the vision transformer indeed provide better performance at a larger adversarial training budget than the baseline. 
An increased weight decay improved the performance only at lower attack budget training, while the networks with SiLU activation performed worse in the closed-loop tests.
At larger attack budgets, no model could drive the car safely, while most models learned by standard ERM could drive all 200 runs flawlessly.

The out-of-training distribution simulation results for autumn and night conditions are shown in the middle and bottom row in Figure \ref{fig:vistaresults}. We observe that adversarial training significantly hurt the out-of-distribution performance of all models, i.e., especially in the autumn data. 
A video demonstration of the simulated runs is available at \url{https://youtu.be/TQKP7l9PfNo}.
In summary, the best driving performance across all four tested conditions was observed with networks trained with standard ERM.

\begin{table}
	\centering
	\begin{tabular}{cl|c|c}\toprule
		Model & Adversarial & Robust & Test \\
		& training budget & validation & accuracy \\
		& & accuracy & \\ \midrule
		\multirow{7}{*}{ResNet50} & $\varepsilon=0$ &  2.3\% $\pm$ 1.9 & \underline{\textbf{93.5\%}} $\pm$ 3.8 \\
		& $\varepsilon=1$  &  18.9\% $\pm$ 3.2 & \textbf{86.1\%} $\pm$ 3.1 \\ 
		& $\varepsilon=2$   &  55.3\% $\pm$ 3.1 & 76.0\% $\pm$ 2.5 \\
		& $\varepsilon=4$    &  77.0\% $\pm$ 3.8 & 68.6\% $\pm$ 3.2 \\
		& $\varepsilon=8$  & 60.7\% $\pm$ 2.7 & 51.8\% $\pm$ 6.6 \\
		& $\varepsilon=4$ (+AWP)  & 75.0\% $\pm$ 1.0  & 67.7\% $\pm$ 8.6 \\ 
		& $\varepsilon=8$ (+AWP)  & 47.6\% $\pm$ 5.5  & 41.4\% $\pm$ 0.4 \\ \midrule
		\multirow{7}{*}{ResNet101} & $\varepsilon=0$  &  3.8\% $\pm$ 1.3  & \textbf{85.9\%} $\pm$ 7.3 \\
		& $\varepsilon=1$ &  20.7\% $\pm$ 2.2 & \textbf{82.9\%} $\pm$ 4.4 \\
		& $\varepsilon=2$  &  54.8\% $\pm$ 1.4 & 76.1\% $\pm$ 2.6 \\ 
		& $\varepsilon=4$ &  44.4\% $\pm$ 0.2 & 41.5\% $\pm$ 1.2 \\
		& $\varepsilon=8$ &  43.9\% $\pm$ 0.5 & 41.7\% $\pm$ 0.0 \\
		& $\varepsilon=4$  (+AWP) & 43.8\% $\pm$ 0.4  & 41.7\% $\pm$ 0.0 \\ 
    	& $\varepsilon=8$ (+AWP)   & 44.2\% $\pm$ 0.4  & 41.7\% $\pm$ 0.0 \\ \midrule
		\multirow{7}{*}{ResNet152} & $\varepsilon=0$ &  4.4\% $\pm$ 4.4 & \textbf{87.2\%} $\pm$ 4.2 \\ 
		& $\varepsilon=1$   &  29.6\% $\pm$ 5.7 & \textbf{82.1\%} $\pm$ 7.0 \\ 
		& $\varepsilon=2$   &  66.7\% $\pm$ 4.2 & 75.8\% $\pm$ 4.4 \\ 
		& $\varepsilon=4$  &  52.6\% $\pm$ 9.8 & 50.8\% $\pm$ 12.4 \\
		& $\varepsilon=8$   &  44.5\% $\pm$ 0.4 & 41.7\% $\pm$ 0.0 \\
		& $\varepsilon=4$  (+AWP)  & \underline{\textbf{78.0\%}} $\pm$ 3.5  & 67.3\% $\pm$ 1.6 \\ 
		& $\varepsilon=8$ (+AWP)  & 44.2\% $\pm$ 0.4  & 41.7\% $\pm$ 0.0 \\ \bottomrule
	\end{tabular}
	\caption{Robust validation accuracy (under I-FGSM with $\varepsilon=8$) and test accuracy on the visual gesture recognition dataset of various adversarially fine-tuned models. Best robust validation accuracy and test accuracies greater than 80\% are highlighted in bold. Best values are underlined.}
	\label{tab:vision}
\end{table}

\begin{table}
	\centering
	\begin{tabular}{cl|c|c}\toprule
		Model & Adversarial & Robust & Test \\
		& training budget & validation & accuracy \\
		& & accuracy & \\ \midrule 
		\multirow{7}{*}{ViT-Small/16} & $\varepsilon=0$  &  0.8\% $\pm$ 1.5 & \textbf{\underline{94.1\%}} $\pm$ 2.1 \\ 
		& $\varepsilon=1$  &  21.3\% $\pm$ 5.6 & \textbf{84.6\%} $\pm$ 6.0 \\
		& $\varepsilon=2$ &  48.0\% $\pm$ 4.8 & \textbf{83.3\%} $\pm$ 4.7 \\
		& $\varepsilon=4$   &  72.3\% $\pm$ 5.0  & 75.2\% $\pm$ 3.1 \\ 
		& $\varepsilon=8$  &  43.9\% $\pm$ 0.4 & 41.1\% $\pm$ 1.9 \\
		& $\varepsilon=4$ (+AWP)  & 58.3\% $\pm$ 2.0  & \textbf{85.5\%} $\pm$ 0.4 \\ 
		  & $\varepsilon=8$ (+AWP)   & 61.4\% $\pm$ 1.8  & 66.6\% $\pm$ 3.7 \\ \midrule
		\multirow{7}{*}{ViT-Base/16} & $\varepsilon=0$ &  11.7\% $\pm$ 4.5 & \textbf{82.3\%} $\pm$ 5.7 \\
		& $\varepsilon=1 $   &  37.9\% $\pm$ 8.4 & 79.4\% $\pm$ 1.7 \\
		& $\varepsilon=2$  &  61.4\% $\pm$ 3.8 & \textbf{84.5\%} $\pm$ 4.6 \\
		& $\varepsilon=4$&  67.1\% $\pm$ 11.7 & 76.0\% $\pm$ 1.9 \\
		& $\varepsilon=8$ &  48.0\% $\pm$ 3.7  & 54.6\% $\pm$ 10.2 \\
		&  $\varepsilon=4$ (+AWP)  & 59.0\% $\pm$ 9.5  & 84.6\% $\pm$ 1.9 \\ 
	    &  $\varepsilon=8$ (+AWP)   & 76.0\% $\pm$ 11.0  & 80.0\% $\pm$ 9.5 \\ \midrule
		\multirow{7}{*}{ViT-Large/16} &  $\varepsilon=0$ &  20.8\% $\pm$ 11.0 & \textbf{85.9\%} $\pm$ 2.5 \\
		& $\varepsilon=1$ &  35.3\% $\pm$ 1.2 & \textbf{89.9\%} $\pm$ 3.6 \\
		& $\varepsilon=2$  &  67.1\% $\pm$ 9.9 & \textbf{89.6\%} $\pm$ 3.2 \\
		& $\varepsilon=4$ &  77.4\% $\pm$ 6.6 & 71.0\% $\pm$ 15.9 \\
		& $\varepsilon=8$  &  58.0\% $\pm$ 17.5 & 47.2\% $\pm$ 9.8 \\
		& $\varepsilon=4$ (+AWP)   & 73.9\% $\pm$ 7.3  & \textbf{86.3\%} $\pm$ 7.4 \\ 
		& $\varepsilon=8$ (+AWP)  & \underline{\textbf{89.7\%}} $\pm$ 0.6  & \textbf{88.9\%} $\pm$ 3.1 \\ \bottomrule
		\end{tabular}
	\caption{Robust validation accuracy (under I-FGSM with $\varepsilon=8$) and test accuracy on the visual gesture recognition dataset of various adversarially fine-tuned models. Best robust validation accuracy and test accuracies greater than 80\% are highlighted in bold. Best values are underlined.}
	\label{tab:vit}
\end{table}

\subsection{Visual gesture recognition}
Our second experiment concerns training an image classifier that controls the operating modes of a mobile robot as reported in \cite{lechner2021adversarial}. The dataset consists of 2029 sample 256-by-256 pixel images corresponding to three classes, i.e., idle (905 samples), enable (552 samples), and disable (572 samples), which are split into a training and a validation set with a 90\%:10\% ratio. The experiments on the physical robot in \cite{lechner2021adversarial} suggest that a validation accuracy of above 90\% is necessary for acceptable robot performance.
Due to the small size of the dataset, we resort to transfer learning of a pre-trained classifier using the big-transfer (BiT) fine-tuning protocol of initializing the output layer with all zeros and training all layers \cite{kolesnikov2020big}. 

In this experiment, we test the theoretical necessity of overparametrization in practice. We train networks of different sizes using adversarial training with increasing attack budget ($\varepsilon \in \{0,1,2,4,8\}$) and report the robust validation accuracy under I-FGSM attacks with a radius of $\varepsilon=8$. We also evaluate models trained with adversarial weight perturbation (AWP) \cite{wu2020adversarial} and $\varepsilon \in \{4,8\}$.

As a proxy for real-world test accuracy, we collect a new dataset comprising 190 idle samples, 129 enable samples, and 140 disable samples. Particularly, the test set resembles a real-world deployment of the model on the robot and ensures that there is no spurious temporal or spatial correlation with the original data source.
We use the clean accuracy of the new set as our test metric to estimate real-world performance.


For increasing the size of the model, we test a ResNet50 (24M), ResNet101 (43M), and ResNet152 (58M) with the number of trainable parameters reported in parenthesis \cite{he2016deep}. 
We also evaluate the vision transformer models ViT-Small (22M), ViT-Base (86M), and ViT-Large (304M) that process the images in the form of 16-by-16 pixel patches \cite{dosovitskiy2020image}.
For the training, we use the Adam optimizer \cite{kingma2014adam} with a learning rate of 0.00005 and a batch size of 64, except for the ResNet152 where a batch size of 32 is used due to out-of-memory errors.  We repeat each training run with 5 random seeds and report the mean and standard deviation.

The results in Table \ref{tab:vision} and Table \ref{tab:vit} show that the overall best test accuracy could be achieved with standard empirical risk minimization and a ResNet50 or ViT-Small model. As expected, however, these models provide no robustness to adversarial attacks. 
Still, acceptable test performance ($\geq$80\%) at non-trivial robustness was realized by models trained with a small attack budget, e.g. $\varepsilon=\{1,2\}$. Nonetheless, the gap between the overall best test accuracy and the top-scoring adversarially trained models is significant, i.e., over one and two standard deviations of the standard trained ResNet50 and ViT-Small model, respectively.

The results in Table \ref{tab:vision} and Table \ref{tab:vit} show the trend that with an increase in model size, the models become more accurate under adversarial training. This effect is even more amplified when considering the more advanced adversarial weight perturbation training (+AWP). Specifically, the most robust ResNet and vision transformer are both their largest variant trained with AWP. Moreover, we observe an advantage of the ViT architecture over the ResNets in terms of robustness, which has been studied in more detail in \cite{paul2022vision,zhou2022understanding}. This result suggests that even larger ViT-based models combined with even more advanced adversarial training schemes may be able to close the robustness-accuracy gap entirely. 

\begin{table*}
	\centering
         \vspace{0.15cm}
	\begin{tabular}{cc|cccc}\toprule
	Safety 	& &  \multicolumn{4}{c}{Validation accuracy} \\
		level & & Width 1  (360k) & Width 2 (1.4M) & Width 3 (3.2M) & Width 4 (5.7M)\\\midrule 
		\multirow{3}{*}{0} & Baseline &  \textbf{83.2\%} $\pm$ 0.8 &  \textbf{84.7\%} $\pm$ 1.6 &  \textbf{83.9\%} $\pm$ 1.9  &  \textbf{85.2\%} $\pm$ 0.8 \\
		 & ELU &   73.3\% $\pm$ 1.5  &  72.5\% $\pm$ 3.3  &  73.3\% $\pm$ 0.8 &  71.3\% $\pm$ 1.3 \\
		& wd+ &  \textbf{82.5\%} $\pm$ 2.0 &  \textbf{84.0\%} $\pm$ 2.1 &  \textbf{85.7\%} $\pm$ 1.3 &  \textbf{85.7}\% $\pm$ 1.2 \\\midrule
	\multirow{3}{*}{1} & Baseline &  75.1\% $\pm$ 2.6 &  78.6\% $\pm$ 3.7 &  77.4\% $\pm$ 2.1   &  78.7\% $\pm$ 3.4 \\
	& ELU  & 53.1\% $\pm$ 0.6 &  53.5\% $\pm$ 0.4 &  52.9\% $\pm$ 0.6 &  52.3\% $\pm$ 0.8  \\
	& wd+  &  74.2\% $\pm$ 3.4 &  75.0\% $\pm$ 1.8 &  65.9\% $\pm$ 10.7 &  67.4\% $\pm$ 12.0  \\\midrule
		\multirow{3}{*}{2} & Baseline &  76.3\% $\pm$ 3.1 &  76.8\% $\pm$ 4.9 &  76.1\% $\pm$ 2.8 &  78.5\% $\pm$ 3.2 \\
		& ELU  &  53.6\% $\pm$ 0.3 &  53.1\% $\pm$ 0.3 &  53.2\% $\pm$ 0.4 &  52.9\% $\pm$ 0.6 \\
		& wd+ &  72.9\% $\pm$ 3.3 &  75.5\% $\pm$ 2.1 &  68.4\% $\pm$ 8.6 &  70.7\% $\pm$ 10.0 \\\midrule
		\multirow{3}{*}{3} & Baseline &  51.8\% $\pm$ 0.9 &  52.8\% $\pm$ 0.5 &  53.3\% $\pm$ 0.1 &  53.9\% $\pm$ 0.3 \\
			& ELU  & 53.2\% $\pm$ 0.8 &  53.8\% $\pm$ 0.5 &  53.1\% $\pm$ 0.1 &  53.2\% $\pm$ 0.4 \\
		& wd+ &  51.4\% $\pm$ 1.1 &  52.8\% $\pm$ 0.7   &  52.8\% $\pm$ 0.6 &  53.4\% $\pm$ 0.4  \\\bottomrule
	\end{tabular}
\caption{Validation accuracy on the robot follow dataset \cite{lechner2021adversarial} of 1D-convolutional NNs with various hyperparameters and trained with standard and safety-domain training. Values greater than 80\% are highlighted in bold. Safety level 0 corresponds to standard training, while the network trained with safety level 1 and above provide formal safety guarantees of never crashing the robot into an obstacle. The columns show networks with different widening factor. The number of learnable parameters are shown in parenthesis. }
\label{tab:follow}
\end{table*}

\subsection{Certified safety-domain training}
Adversarial training methods do not ensure robustness but provide only empirical improvements over common attack methods. 
Certified training methods such as the interval bound propagation \cite{gowal2019scalable} can learn networks with formal robustness or safety guarantees.
In this experiment, we study the safety-domain training of LiDAR-based mobile robot navigation controller \cite{lechner2021adversarial}. The objective of the learned controller is to map 541-dimensional laser range scans to 7 possible categories, i.e., stay, straight forward, left forward, right forward, straight backward, left backward, and right backward. The dataset consists of 2705 training and 570 validation samples uniformly distributed across the seven classes. 
Using safety-domain training, we want to ensure that the robot never crashes into an object in front of it. This is achieved by training an abstract interpretation representation of the network to never output a forward locomotion class in case the LiDAR input indicates an obstacle.
There are four safety levels with different strictness of what accounts for an obstacle, e.g., several consecutive rays or just a single ray, defined in \cite{lechner2021adversarial}. Safety level 0 corresponds to standard training, while safety level 3 represents the strictest level.

We test the overparametrization, increased weight decay (from $0$ to $10^{-5}$), and smooth activation function methods on this task. As a baseline, we use the 1D-CNN from \cite{lechner2021adversarial}, which is comprised of 360k parameters. Our overparametrized models increase the width of the network to obtain CNNs with 1.4M, 3.2M, and 5.7M parameters respectively. We use the exponential linear unit (ELU) activation function \cite{clevert2015fast} to represent a smooth activation due to the non-monotonicity of SiLU being incompatible with the used abstract interpretation domains. 
We train all models with the Adam optimizer \cite{kingma2014adam} with a learning rate of 0.0001 and a batch size of 64. The safety level 0 models are trained for 20 epochs, while the networks trained using safety-domain training for 2000 epochs. The network architectures are shown in the Appendix.

We report the validation accuracy as an evaluation metric. The experiments on the physical robot in \cite{lechner2021adversarial} suggest that a validation accuracy above 80\% is necessary to achieve an acceptable real-world performance.  Note that all models, except those trained with safety level 0, provide some form of formal safety guarantees. Therefore, this experiment studies how much validation accuracy is traded for the ensured safety.  We repeat each training run with 5 random seeds and report the mean and standard deviation.

The result in Table \ref{tab:follow} shows that safety-domain training benefits from an increased number of parameters (width). However, the improvement over the baseline is rather incremental and accounts only for a few percent. In contrast, the accuracy reduction caused by the safety-domain training is several times more significant, e.g., around 10\%, and no network trained with safety-domain training exceeds the threshold of 80\% accuracy.
The networks with smooth activation function and increased weight decay performed worse than the baseline when using safety-domain training. This suggests that certified training methods such as safety-domain training may require different hyperparameters and learning settings than adversarial training.

\section{Discussion and Conclusion}\label{sec:conclusion}
Adversarial training (i.e., training on adversarially perturbed input data) is a well-studied method for making neural networks robust to potential adversarial attacks during inference.
However, the improved robustness does not come for free but rather is accompanied by a decrease in nominal model accuracy and performance \cite{zhang2019theoretically}.
Recent work \cite{lechner2021adversarial} has shown that, in practical robot learning applications, the effects of adversarial training do not pose a fair trade-off but inflict a net loss when measured in holistic robot performance.
This work revisited the robustness-accuracy trade-off in robot learning by systematically analyzing if recent advances in robust training methods and theory in conjunction with adversarial robot learning can make adversarial training suitable for real-world robot applications.

We evaluated a total of five robust training methods on three different robot learning tasks ranging from autonomous driving in a high-fidelity environment amenable to sim-to-real deployment to mobile robot navigation and gesture recognition. Our results indicate that the negative impact on the nominal accuracy from adversarial training still outweighs the induced robustness. In other words, while adversarial training can improve the model's ability to withstand attacks, it does not justify the reduced accuracy on clean, non-adversarial data. 

Nonetheless, our results suggest that, in aggregate, when combining these methods, a significant improvement in the robustness-accuracy gap is made. For instance, the combination of overparametrization, a vision transformer, and a more advanced training scheme (adversarial weight perturbation) performs much better under adversarial training than the models tested in \cite{lechner2021adversarial}. This suggests that future research directions that can be further combined, e.g., data augmentation or other training schemes, may be able to close the robustness-accuracy entirely.
%
\section*{Acknowledgment} \label{section: acknowledgement}
This work was supported in parts by the AI2050 program at Schmidt Futures (Grant G-22-63172), Capgemini SE, ERC-2020-AdG 101020093, National Science Foundation (NSF), and JP Morgan Graduate Fellowships. We thank Christoph Lampert for inspiring this work.
Research was sponsored by the United States Air Force Research Laboratory and the United States Air Force Artificial Intelligence Accelerator and was accomplished under Cooperative Agreement Number FA8750-19-2-1000. The views and conclusions contained in this document are those of the authors and should not be interpreted as representing the official policies, either expressed or implied, of the United States Air Force or the U.S. Government. The U.S. Government is authorized to reproduce and distribute reprints for Government purposes notwithstanding any copyright notation herein.

\bibliographystyle{IEEEtran}
\bibliography{IEEEabrv,references}

\begin{thebibliography}{10}
\providecommand{\url}[1]{#1}
\csname url@samestyle\endcsname
\providecommand{\newblock}{\relax}
\providecommand{\bibinfo}[2]{#2}
\providecommand{\BIBentrySTDinterwordspacing}{\spaceskip=0pt\relax}
\providecommand{\BIBentryALTinterwordstretchfactor}{4}
\providecommand{\BIBentryALTinterwordspacing}{\spaceskip=\fontdimen2\font plus
\BIBentryALTinterwordstretchfactor\fontdimen3\font minus
  \fontdimen4\font\relax}
\providecommand{\BIBforeignlanguage}[2]{{%
\expandafter\ifx\csname l@#1\endcsname\relax
\typeout{** WARNING: IEEEtran.bst: No hyphenation pattern has been}%
\typeout{** loaded for the language `#1'. Using the pattern for}%
\typeout{** the default language instead.}%
\else
\language=\csname l@#1\endcsname
\fi
#2}}
\providecommand{\BIBdecl}{\relax}
\BIBdecl

\bibitem{szegedy2013intriguing}
C.~Szegedy, W.~Zaremba, I.~Sutskever, J.~Bruna, D.~Erhan, I.~Goodfellow, and
  R.~Fergus, ``Intriguing properties of neural networks,'' \emph{arXiv preprint
  arXiv:1312.6199}, 2013.

\bibitem{goodfellow2014explaining}
I.~J. Goodfellow, J.~Shlens, and C.~Szegedy, ``Explaining and harnessing
  adversarial examples,'' \emph{arXiv preprint arXiv:1412.6572}, 2014.

\bibitem{wald1945statistical}
A.~Wald, ``Statistical decision functions which minimize the maximum risk,''
  \emph{Annals of Mathematics}, pp. 265--280, 1945.

\bibitem{huber1964robust}
P.~J. Huber, ``Robust estimation of a location parameter,'' \emph{The Annals of
  Mathematical Statistics}, vol.~1, no.~35, 1964.

\bibitem{atkeson1997robot}
C.~G. Atkeson and S.~Schaal, ``Robot learning from demonstration,'' in
  \emph{International Conference on Machine Learning (ICML)}, 1997.

\bibitem{xu2009robustness}
H.~Xu, C.~Caramanis, and S.~Mannor, ``Robustness and regularization of support
  vector machines.'' \emph{The Journal of Machine Learning Research (JMLR)},
  vol.~10, 2009.

\bibitem{madry2017towards}
A.~Madry, A.~Makelov, L.~Schmidt, D.~Tsipras, and A.~Vladu, ``Towards deep
  learning models resistant to adversarial attacks,'' \emph{arXiv preprint
  arXiv:1706.06083}, 2017.

\bibitem{song2018improving}
C.~Song, K.~He, L.~Wang, and J.~E. Hopcroft, ``Improving the generalization of
  adversarial training with domain adaptation,'' in \emph{International
  Conference on Learning Representations (ICLR)}, 2019.

\bibitem{biggio2012poisoning}
B.~Biggio, B.~Nelson, and P.~Laskov, ``Poisoning attacks against support vector
  machines,'' in \emph{International Conference on Machine Learning (ICML)},
  2012.

\bibitem{konstantinov2019robust}
N.~Konstantinov and C.~Lampert, ``Robust learning from untrusted sources,'' in
  \emph{International Conference on Machine Learning (ICML)}, 2019.

\bibitem{raghunathan2019adversarial}
A.~Raghunathan, S.~M. Xie, F.~Yang, J.~C. Duchi, and P.~Liang, ``Adversarial
  training can hurt generalization,'' \emph{arXiv preprint arXiv:1906.06032},
  2019.

\bibitem{zhang2018efficient}
H.~Zhang, T.-W. Weng, P.-Y. Chen, C.-J. Hsieh, and L.~Daniel, ``Efficient
  neural network robustness certification with general activation functions,''
  in \emph{Conference on Neural Information Processing Systems (NeurIPS)},
  2018.

\bibitem{tsipras2018robustness}
D.~Tsipras, S.~Santurkar, L.~Engstrom, A.~Turner, and A.~Madry, ``Robustness
  may be at odds with accuracy,'' in \emph{International Conference on Learning
  Representations (ICLR)}, 2018.

\bibitem{zhang2019theoretically}
H.~Zhang, Y.~Yu, J.~Jiao, E.~Xing, L.~El~Ghaoui, and M.~Jordan, ``Theoretically
  principled trade-off between robustness and accuracy,'' in
  \emph{International Conference on Machine Learning (ICML)}, 2019.

\bibitem{zagoruyko2016wide}
S.~Zagoruyko and N.~Komodakis, ``Wide residual networks,'' in \emph{British
  Machine Vision Conference (BMVC)}, 2016.

\bibitem{lechner2021adversarial}
M.~Lechner, R.~M. Hasani, R.~Grosu, D.~Rus, and T.~A. Henzinger, ``Adversarial
  training is not ready for robot learning,'' in \emph{IEEE International
  Conference on Robotics and Automation (ICRA)}, 2021.

\bibitem{rebuffi2021data}
S.-A. Rebuffi, S.~Gowal, D.~A. Calian, F.~Stimberg, O.~Wiles, and T.~A. Mann,
  ``Data augmentation can improve robustness,'' in \emph{Conference on Neural
  Information Processing Systems (NeurIPS)}, 2021.

\bibitem{bubeck2021law}
S.~Bubeck, Y.~Li, and D.~M. Nagaraj, ``A law of robustness for two-layers
  neural networks,'' in \emph{Conference on Learning Theory (COLT)}, 2021.

\bibitem{zhou2022understanding}
D.~Zhou, Z.~Yu, E.~Xie, C.~Xiao, A.~Anandkumar, J.~Feng, and J.~M. Alvarez,
  ``Understanding the robustness in vision transformers,'' in
  \emph{International Conference on Machine Learning (ICML)}, 2022, pp.
  27\,378--27\,394.

\bibitem{singla2021low}
V.~Singla, S.~Singla, S.~Feizi, and D.~Jacobs, ``Low curvature activations
  reduce overfitting in adversarial training,'' in \emph{IEEE International
  Conference on Computer Vision (ICCV)}, 2021.

\bibitem{pang2021bag}
T.~Pang, X.~Yang, Y.~Dong, H.~Su, and J.~Zhu, ``Bag of tricks for adversarial
  training,'' in \emph{International Conference on Learning Representations
  (ICLR)}, 2021.

\bibitem{wu2020adversarial}
D.~Wu, S.-T. Xia, and Y.~Wang, ``Adversarial weight perturbation helps robust
  generalization,'' in \emph{Conference on Neural Information Processing
  Systems (NeurIPS)}, 2020.

\bibitem{bubeck2021universal}
S.~Bubeck and M.~Sellke, ``A universal law of robustness via isoperimetry,'' in
  \emph{Conference on Neural Information Processing Systems (NeurIPS)}, 2021.

\bibitem{paul2022vision}
S.~Paul and P.-Y. Chen, ``Vision transformers are robust learners,'' in
  \emph{AAAI Conference on Artificial Intelligence (AAAI)}, vol.~36, no.~2,
  2022, pp. 2071--2081.

\bibitem{brown2017adversarial}
T.~B. Brown, D.~Man{\'e}, A.~Roy, M.~Abadi, and J.~Gilmer, ``Adversarial
  patch,'' in \emph{Conference on Neural Information Processing Systems
  (NeurIPS)}, 2017.

\bibitem{uesato2018adversarial}
J.~Uesato, B.~O'Donoghue, A.~v.~d. Oord, and P.~Kohli, ``Adversarial risk and
  the dangers of evaluating against weak attacks,'' in \emph{International
  Conference on Machine Learning (ICML)}, 2018.

\bibitem{athalye2018obfuscated}
A.~Athalye, N.~Carlini, and D.~Wagner, ``Obfuscated gradients give a false
  sense of security: Circumventing defenses to adversarial examples,''
  \emph{arXiv preprint arXiv:1802.00420}, 2018.

\bibitem{schonherr2018adversarial}
L.~Sch{\"o}nherr, K.~Kohls, S.~Zeiler, T.~Holz, and D.~Kolossa, ``Adversarial
  attacks against automatic speech recognition systems via psychoacoustic
  hiding,'' \emph{arXiv preprint arXiv:1808.05665}, 2018.

\bibitem{giacobbe2020many}
M.~Giacobbe, T.~A. Henzinger, and M.~Lechner, ``How many bits does it take to
  quantize your neural network?'' in \emph{International Conference on Tools
  and Algorithms for the Construction and Analysis of Systems (TACAS)}, 2020.

\bibitem{bai2021transformers}
Y.~Bai, J.~Mei, A.~L. Yuille, and C.~Xie, ``Are transformers more robust than
  cnns?'' in \emph{Conference on Neural Information Processing Systems
  (NeurIPS)}, 2021.

\bibitem{katz2017reluplex}
G.~Katz, C.~Barrett, D.~L. Dill, K.~Julian, and M.~J. Kochenderfer, ``Reluplex:
  An efficient smt solver for verifying deep neural networks,'' in
  \emph{International Conference on Computer Aided Verification (CAV)}, 2017.

\bibitem{salzer2021reachability}
M.~S{\"a}lzer and M.~Lange, ``Reachability is np-complete even for the simplest
  neural networks,'' in \emph{International Conference on Reachability
  Problems}, 2021.

\bibitem{henzinger2021scalable}
T.~A. Henzinger, M.~Lechner, and {\DJ}.~{\v{Z}}ikeli{\'c}, ``Scalable
  verification of quantized neural networks,'' in \emph{AAAI Conference on
  Artificial Intelligence (AAAI)}, 2021.

\bibitem{kurakin2016adversarial}
A.~Kurakin, I.~Goodfellow, and S.~Bengio, ``Adversarial machine learning at
  scale,'' in \emph{International Conference on Learning Representations
  (ICLR)}, 2017.

\bibitem{moosavi2016deepfool}
S.-M. Moosavi-Dezfooli, A.~Fawzi, and P.~Frossard, ``Deepfool: a simple and
  accurate method to fool deep neural networks,'' in \emph{IEEE Conference on
  Computer Vision and Pattern Recognition (CVPR)}, 2016.

\bibitem{carlini2017towards}
N.~Carlini and D.~Wagner, ``Towards evaluating the robustness of neural
  networks,'' in \emph{IEEE Symposium on Security and Privacy}, 2017.

\bibitem{shafahi2019forfree}
A.~Shafahi, M.~Najibi, M.~A. Ghiasi, Z.~Xu, J.~Dickerson, C.~Studer, L.~S.
  Davis, G.~Taylor, and T.~Goldstein, ``Adversarial training for free!'' in
  \emph{Conference on Neural Information Processing Systems (NeurIPS)}, 2019.

\bibitem{lamb2022interpolated}
A.~Lamb, V.~Verma, K.~Kawaguchi, A.~Matyasko, S.~Khosla, J.~Kannala, and
  Y.~Bengio, ``Interpolated adversarial training: Achieving robust neural
  networks without sacrificing too much accuracy,'' \emph{Neural Networks},
  vol. 154, pp. 218--233, 2022.

\bibitem{zhang2018mixup}
\BIBentryALTinterwordspacing
H.~Zhang, M.~Cisse, Y.~N. Dauphin, and D.~Lopez-Paz, ``mixup: Beyond empirical
  risk minimization,'' in \emph{International Conference on Learning
  Representations}, 2018. [Online]. Available:
  \url{https://openreview.net/forum?id=r1Ddp1-Rb}
\BIBentrySTDinterwordspacing

\bibitem{ChangRG19}
Y.~Chang, N.~Roohi, and S.~Gao, ``Neural lyapunov control,'' in
  \emph{Conference on Neural Information Processing Systems (NeurIPS)}, 2019.

\bibitem{lechner2021infinite}
M.~Lechner, D.~Zikelic, K.~Chatterjee, and T.~A. Henzinger, ``Infinite time
  horizon safety of bayesian neural networks,'' in \emph{Conference on Neural
  Information Processing Systems (NeurIPS)}, 2021.

\bibitem{lechner2021stability}
M.~Lechner, {\DJ}.~{\v{Z}}ikeli{\'c}, K.~Chatterjee, and T.~A. Henzinger,
  ``Stability verification in stochastic control systems via neural network
  supermartingales,'' in \emph{AAAI Conference on Artificial Intelligence
  (AAAI)}, 2022.

\bibitem{herrmann2022pyramid}
C.~Herrmann, K.~Sargent, L.~Jiang, R.~Zabih, H.~Chang, C.~Liu, D.~Krishnan, and
  D.~Sun, ``Pyramid adversarial training improves vit performance,'' in
  \emph{Proceedings of the IEEE/CVF Conference on Computer Vision and Pattern
  Recognition}, 2022, pp. 13\,419--13\,429.

\bibitem{dosovitskiy2020image}
A.~Dosovitskiy, L.~Beyer, A.~Kolesnikov, D.~Weissenborn, X.~Zhai,
  T.~Unterthiner, M.~Dehghani, M.~Minderer, G.~Heigold, S.~Gelly \emph{et~al.},
  ``An image is worth 16x16 words: Transformers for image recognition at
  scale,'' in \emph{International Conference on Learning Representations
  (ICLR)}, 2020.

\bibitem{duan2019robot}
J.~Duan, Q.~Wang, L.~Pinto, C.-C.~J. Kuo, and S.~Nikolaidis, ``Robot learning
  via human adversarial games,'' in \emph{2019 IEEE/RSJ International
  Conference on Intelligent Robots and Systems (IROS)}.\hskip 1em plus 0.5em
  minus 0.4em\relax IEEE, 2019, pp. 1056--1063.

\bibitem{elfwing2018sigmoid}
S.~Elfwing, E.~Uchibe, and K.~Doya, ``Sigmoid-weighted linear units for neural
  network function approximation in reinforcement learning,'' \emph{Neural
  Networks}, vol. 107, pp. 3--11, 2018.

\bibitem{ramachandran2017searching}
P.~Ramachandran, B.~Zoph, and Q.~V. Le, ``Searching for activation functions,''
  \emph{arXiv preprint arXiv:1710.05941}, 2017.

\bibitem{amini2021vista}
A.~Amini, T.-H. Wang, I.~Gilitschenski, W.~Schwarting, Z.~Liu, S.~Han,
  S.~Karaman, and D.~Rus, ``Vista 2.0: An open, data-driven simulator for
  multimodal sensing and policy learning for autonomous vehicles,'' \emph{arXiv
  preprint arXiv:2111.12083}, 2021.

\bibitem{vaswani2017attention}
A.~Vaswani, N.~Shazeer, N.~Parmar, J.~Uszkoreit, L.~Jones, A.~N. Gomez,
  {\L}.~Kaiser, and I.~Polosukhin, ``Attention is all you need,'' in
  \emph{Conference on Neural Information Processing Systems (NeurIPS)}, 2017.

\bibitem{naseer2021intriguing}
M.~Naseer, K.~Ranasinghe, S.~Khan, M.~Hayat, F.~S. Khan, and M.-H. Yang,
  ``Intriguing properties of vision transformers,'' in \emph{Conference on
  Neural Information Processing Systems (NeurIPS)}, 2021.

\bibitem{kingma2014adam}
D.~P. Kingma and J.~Ba, ``Adam: A method for stochastic optimization,''
  \emph{arXiv preprint arXiv:1412.6980}, 2014.

\bibitem{kolesnikov2020big}
A.~Kolesnikov, L.~Beyer, X.~Zhai, J.~Puigcerver, J.~Yung, S.~Gelly, and
  N.~Houlsby, ``Big transfer (bit): General visual representation learning,''
  in \emph{European Conference on Computer Vision (ECCV)}, 2020.

\bibitem{he2016deep}
K.~He, X.~Zhang, S.~Ren, and J.~Sun, ``Deep residual learning for image
  recognition,'' in \emph{IEEE Conference on Computer Vision and Pattern
  Recognition (CVPR)}, 2016.

\bibitem{gowal2019scalable}
S.~Gowal, K.~D. Dvijotham, R.~Stanforth, R.~Bunel, C.~Qin, J.~Uesato,
  R.~Arandjelovic, T.~Mann, and P.~Kohli, ``Scalable verified training for
  provably robust image classification,'' in \emph{IEEE International
  Conference on Computer Vision (ICCV)}, 2019.

\bibitem{clevert2015fast}
D.-A. Clevert, T.~Unterthiner, and S.~Hochreiter, ``Fast and accurate deep
  network learning by exponential linear units (elus),'' \emph{arXiv preprint
  arXiv:1511.07289}, 2015.

\end{thebibliography}

\newpage
\newpage
\thispagestyle{empty}

\appendix
\section{Network architecture details}\label{app:details}
Here we describe the details of the neural network architectures used in our experiments.

\begin{table}[b]
	\centering
	\begin{tabular}{c|c}\toprule
		Layer & Parameter \\\midrule
		Conv2D & F=32, K=5, S=2, ReLU \\
		Conv2D & F=64, K=5, S=1, ReLU \\
		Conv2D & F=96, K=3, S=2, ReLU \\
		Conv2D & F=128, K=3, S=1, ReLU \\
		GlobalAveragePool2D & \\
		Fully-connected & 1000 units, ReLU  \\
		Fully-connected & 100 units, ReLU  \\
		Fully-connected & 1 unit \\\bottomrule
	\end{tabular}
\caption{Convolutional neural network baseline architecture for our autonomous driving experiment (440k parameters). F refers to the number of filters, K to the kernel size, and S to the stride.}
\label{tab:vistacnn}
\end{table}

\begin{table}[b]
	\centering
	\begin{tabular}{c|c}\toprule
		Layer & Parameter \\\midrule
		Conv2D & F=32, K=5, S=2\\
		BatchNorm2D & ReLU (post BN)\\
		Conv2D & F=128, K=5, S=1\\
		BatchNorm2D & ReLU (post BN)\\
		Conv2D & F=256, K=3, S=2\\
		BatchNorm2D & ReLU (post BN)\\
		Conv2D & F=512, K=3, S=1\\
		BatchNorm2D & ReLU (post BN)\\
		Conv2D & F=1024, K=3, S=1\\
		GlobalAveragePool2D & \\
		Fully-connected & 1024 units, ReLU \\
		Fully-connected & 256 units, ReLU \\
		Fully-connected & 1 unit \\\bottomrule
	\end{tabular}
	\caption{Enlarged neural network architecture for our autonomous driving experiment (7.7M parameters). F refers to the number of filters, K to the kernel size, and S to the stride. }
	\label{tab:vistabigcnn}
\end{table}

\begin{table}[b]
	\centering
	\begin{tabular}{c|c}\toprule
		Layer & Parameter \\\midrule
		Conv1D & F=w*32, K=5, S=1, ReLU \\
		Conv1D & F=w*96, K=5, S=2, ReLU \\
		Conv1D & F=w*96, K=5, S=2, ReLU \\
		Conv1D & F=w*96, K=5, S=2, ReLU \\
		Conv1D & F=w*96, K=5, S=2, ReLU \\
		Conv1D & F=w*96, K=5, S=2, ReLU \\
		Flatten & \\
		Fully-connected & w*128 units, ReLU \\
		Fully-connected 7 & softmax\\\bottomrule
	\end{tabular}
\caption{Network architecture of the 1D-CNN trained with safety-domain training (F= number of filters, K = kernel size, S = stride). w is the widening factor. }
\label{tab:follownetwork}
\vspace{-15pt}
\end{table}

\end{document}